\documentclass{article}
\usepackage[eatrack]{log_2023}

\usepackage{booktabs}            
\usepackage{multirow}            
\usepackage{amsfonts}            
\usepackage{amssymb}
\usepackage{graphicx}            
\usepackage{amsthm}			     
\usepackage{siunitx}
\usepackage{nicefrac}
\usepackage{textcomp}
\usepackage{xcolor}

\usepackage{tikz}
\PassOptionsToPackage{RGB}{xcolor}

\definecolor{custom3}{RGB}{2, 158, 115}
\definecolor{custom4}{RGB}{213, 94, 0}

\newcommand{\tikzxmark}{%
\tikz[scale=0.23,custom4] {
    \draw[line width=1.3,line cap=round] (0,0) to [bend left=6] (1,1);
    \draw[line width=1.3,line cap=round] (0.2,0.95) to [bend right=3] (0.8,0.05);
}}
\newcommand{\tikzcmark}{%
\tikz[scale=0.23,custom3] {
    \draw[line width=1.3,line cap=round] (0.25,0) to [bend left=10] (1,1);
    \draw[line width=1.5,line cap=round] (0,0.35) to [bend right=1] (0.23,0);
}}

\definecolor{custom2}{RGB}{222, 143, 5}

\newtheorem{lemma}{Lemma}

\usepackage[
     backend=biber,
     style=numeric-comp,
     backref=true,
     natbib=true]{biblatex}
\title[The Self-Loop Paradox: Investigating the Impact of Self-loops on GNNs]{The Self-Loop Paradox: Investigating the Impact of Self-Loops on Graph Neural Networks}

\author[M. Lampert et al.]{%
Moritz Lampert, Ingo Scholtes\\
Chair of Machine Learning for Complex Networks\\
Center for Artificial Intelligence and Data Science (CAIDAS)\\
Julius-Maximilians-Universit\"at W\"urzburg, DE\\
\email{moritz.lampert@uni-wuerzburg.de}
}

\begin{document}

\maketitle

\begin{abstract}
	Many Graph Neural Networks (GNNs) add self-loops to a graph to include feature information about a node itself at each layer.
	However, if the GNN consists of more than one layer, this information can return to its origin via cycles in the graph topology.
	Intuition suggests that this ``backflow'' of information should be larger in graphs with self-loops compared to graphs without.
	In this work, we counter this intuition and show that for certain GNN architectures, the information a node gains from itself can be smaller in graphs with self-loops compared to the same graphs without.
	We adopt an analytical approach for the study of statistical graph ensembles with a given degree sequence and show that this phenomenon, which we call the \emph{self-loop paradox}, can depend both on the number of GNN layers $k$ and whether $k$ is even or odd.
	We experimentally validate our theoretical findings in a synthetic node classification task and investigate its practical relevance in 23 real-world graphs.
\end{abstract}

\section{Introduction}

Multi-layer Perceptrons (MLPs) use feature information of a sample to make predictions. 
GNNs use the sample represented as a node in a graph and feature information of the neighbouring nodes for predictions. 
To include the information about a node itself, GNN architectures like Graph Convolutional Network (GCN) \cite{SemiSupervisedClassificationGraph2017kipf} add self-loops.
For single-layer GNNs, it is necessary to include the node's feature but for more than one layer, information can also flow back via cycles in the topology of the graph. 
Intuitively, adding self-loops still increases the information a node learns about itself.
Thus, the influence of self-loops on GNNs remains mainly unexplored up to now. \\
Recent works investigate the influence of a graph's topology on GNNs in combination with over-squashing, i.e.\ GNNs perform badly on tasks that rely on interactions between distant nodes.
The exponentially growing number of neighbours with increasing distance is one reason for over-squashing \cite{BottleneckGraphNeural2021alon}.
The GNNs perform badly because useful information of distant nodes is squashed by bottlenecks in the graph topology \cite{OverSquashingMessagePassing2023giovannia,UnderstandingOversquashingBottlenecks2022topping} and the redundancy introduced by close neighbours \cite{RedundancyFreeMessagePassing2022chen}. 
The works on over-squashing all have in common that they connect the graph topology via walks between nodes to the influence nodes have on each other's predictions in a GNN.
Earlier work \cite{RepresentationLearningGraphs2018xub} uses a similar approach in a different context and also mentions how self-loops can influence GNN predictions. 
\citet{UnderstandingOversquashingBottlenecks2022topping} build on that idea and investigate it further. \\
We build upon the concept that the influence of a node feature on the prediction of another node is proportional to the relative number of walks between them out of all walks. 
In particular, we investigate the proportion of cycles of length $k$ from a node $v$ out of all walks of length $k$ ending in $v$. 
Using theory from network science based on random graphs with a given degree sequence \cite{RandomGraphsArbitrary2001newman,VertexSimilarityNetworks2006leicht}, we prove that including self-loops in those random graphs decreases the proportion of cycles with length $2$. We further show that this can reduce a node's influence on its own prediction in two-layered GNNs.
The theoretical findings are validated empirically using synthetic data generated using a Stochastic Block Model (SBM) \cite{StochasticBlockmodelsFirst1983holland} and put into context on real-world datasets in the appendix. 

\section{Background}

\paragraph{Random Graphs} For the statistical analysis, we adopt the probabilistic Molloy-Reed configuration model \cite{CriticalPointRandom1995molloy} that generates random undirected graphs $\mathcal{G}=(\mathbf{V}, \mathbf{E})$ based on a given degree sequence $S = (d_v)_{v\in \mathbf{V}}$.
Under this model, if the degree sequence $S$ is graphic, each graph $\mathcal{G}$ with the same degree sequence $S$ is equiprobable.
Computationally, this can be achieved by adding $d_v$ edge stubs for each node $v$ with degree $d_v$ and then connecting pairs of randomly chosen stubs until no stubs are left. \\
While it is easy to algorithmically generate random graphs based on the Molloy-Reed model, its true importance in network science is due to our ability to \emph{analytically} study expected properties of these random graphs based on generating functions \cite{wilf2005generatingfunctionology,RandomGraphsArbitrary2001newman}.
For this type of model, we can express the expected degree of a randomly chosen node as%
\begin{align}%
    \langle d \rangle = \frac{1}{|\mathbf{V}|} \sum_{d_v \in \mathbf{S}} d_v.
\end{align}%
The expected degree of a random neighbour of a randomly chosen node is given by the expected degree and its second raw moment: \cite{RandomGraphsArbitrary2001newman}%
\begin{align}%
    \langle d_N \rangle = \frac{\langle d^2 \rangle}{\langle d \rangle}
\end{align}%
The number of walks of length $k$ from node $u$ to $v$ can be computed using the adjacency matrix $A$ of graph $\mathcal{G}$ as $A^k_{uv}$.
The expected total number of $k$-length walks from or to $v$ is given by \cite{RandomGraphsArbitrary2001newman}%
\begin{align}%
    \mathbb{E}\left(\sum_{u \in \mathbf{V}} A^k_{uv}\right) = \mathbb{E}\left(\sum_{u \in \mathbf{V}} A^k_{vu}\right) = \langle d \rangle \cdot \langle d_N \rangle^{k-1} = \langle d \rangle \cdot \left(\frac{\langle d^2 \rangle}{\langle d \rangle}\right)^{k-1}.
\end{align}%

\paragraph{Graph Neural Networks} Let $\mathcal{G}= (\mathbf{V}, \mathbf{E})$ be a graph without self-loops, with $n$ nodes, $m$ edges and adjacency matrix $A$. 
Then given the input features of the previous layer $\mathbf{h}_u^{(k-1)}$, a GCN-layer \cite{SemiSupervisedClassificationGraph2017kipf} is defined as%
\begin{align}\label{equ:gcn layer}%
    \mathbf{h}_v^{(k)} =  \sigma^{(k)}\left(\left(\sum_{u \in N(v) \cup \{v\}} \frac{1}{\sqrt{d(v)d(u)}} \cdot \mathbf{h}_u^{(k-1)}\right) \cdot W^{(k)} + b^{(k)} \right),
\end{align}%
where $N(v)$ yields the set of neighbours of $v$ and $d(v)$ is the degree of $v$.
The weight matrix $W^{(k)}$ and the bias vector $b^{(k)}$ are trainable parameters of the $k$-th layer and $\sigma^{(k)}$ is a non-linear activation function. \\
Note that the feature $\mathbf{h}_v^{(k)}$ of the node $v$ itself is included in the summation without any special treatment. 
This is the same as adding self-loops to the graph topology and using the adjacency matrix $\tilde{A} = A + I$.
Other GNNs like GraphSAGE \cite{InductiveRepresentationLearning2017hamilton} or the Graph Isomorphism Network (GIN) \cite{HowPowerfulAre2019xu} treat the node's feature differently and, thus, the theoretical findings of this work do not directly apply to these GNNs (see Appendix \ref{app:gnn types}). \\
To measure the influence of an input feature on the prediction of a node $v$, we use the Jacobian of the output with respect to a certain input feature and make use of a finding by \citet{RedundancyFreeMessagePassing2022chen} in the context of over-squashing for Message Passing GNNs (MPNNs) \cite{NeuralMessagePassing2017gilmer}:%
\begin{lemma}\label{lem:influence of input feature}%
    If the MPNN passes messages along all $k$-length walks from $u$ to $v$ with equal probability, then the relative influence of input feature $\mathbf{h}_u^{(0)}$ on node output $\mathbf{h}_v^{(k)}$ is on average%
    \begin{align}%
        \mathbb{E}\left(\frac{\partial \mathbf{h}_v^{(k)}/\partial \mathbf{h}_u^{(0)}}{\sum_{u'\in V} \partial \mathbf{h}_v^{(k)}/\partial \mathbf{h}_{u'}^{(0)}}\right) = \frac{\tilde{A}^k_{uv}}{\sum_{u'\in V} \tilde{A}^k_{u'v}}.
    \end{align}%
\end{lemma}%
This means that it is possible to make statements about the influence of specific input features on the prediction by analysing the graph topology and the number of walks of a certain length between two nodes.

\section{The Self-Loop Paradox}
\paragraph{Theoretical Analysis} We prove that relatively speaking, more walks lead back to the node itself in graphs without self-loops compared to graphs with self-loops:%
\begin{lemma}\label{lem:self-loop paradox}%
    Given a graph $\mathcal{G}$ generated using the configuration model \cite{CriticalPointRandom1995molloy} with adjacency matrix $A$ without self-loops and its counterpart with self-loops $\tilde{\mathcal{G}}$ with $\tilde{A}=A+I$, the proportion of cycles of length $2$ from a node $v$ to itself out of all walks of length $2$ ending in $v$ is larger in $\mathcal{G}$ than in $\tilde{\mathcal{G}}$:%
    \begin{align}%
        \frac{\mathbb{E}\left(A^2_{vv}\right)}{\mathbb{E}\left(\sum_{u \in V} A^2_{uv}\right)} > \frac{\mathbb{E}\left(\tilde{A}^2_{vv}\right)}{\mathbb{E}\left(\sum_{u \in V} \tilde{A}^2_{uv}\right)}.
    \end{align}%
\end{lemma}%
We prove Lemma \ref{lem:self-loop paradox} in Appendix \ref{sec:lemma2proof} and experimentally validate the given proportions on example graphs in Appendix \ref{app:walk statistics}.
This means in combination with Lemma \ref{lem:influence of input feature} that including self-loops can decrease the information a node retains about itself in certain MPNNs with two layers. A detailed discussion on which GNNs are affected is provided in Appendix \ref{app:gnn types}.
This is counter-intuitive since for single-layer GNNs, the node's feature is only included via self-loops. Although strictly speaking, this only applies to graphs generated from the configuration model, we show that this effect can also be observed in other random graphs below and real-world networks in Appendix C. This is in line with other properties of the configuration model that are also observed in real-world networks \cite[377--381]{Networks2018newman}.

\begin{figure}[!ht]%
    \centering%
    \includegraphics{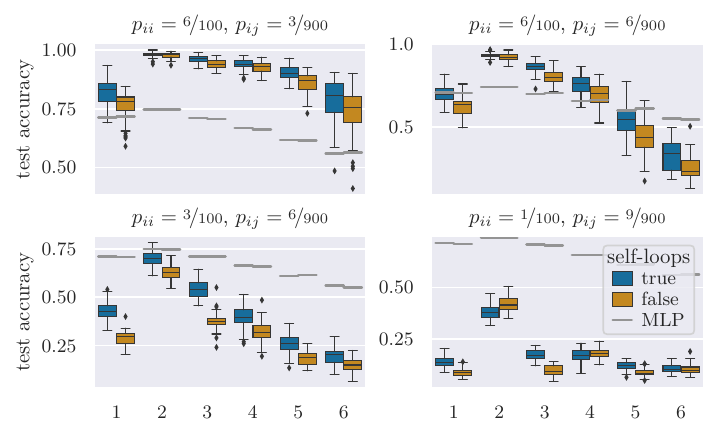}%
    \caption{Test set accuracies of GCNs with and without self-loops averaged over 50 runs for different numbers of layers. Coloured boxes mark the range between the first and third quartiles. Whiskers stretch to the furthest outliers within three halves of that range while other outliers outside of the whiskers are marked with a rhombus. Grey lines show average accuracies of MLPs with the corresponding number of layers.}\label{fig:random evaluation}%
\end{figure}

\paragraph{Empirical Validation} To empirically validate our theoretical results on GNNs, we use a GCN model to address a node classification task in synthetic graphs generated based on the Stochastic Block Model (SBM) \cite{StochasticBlockmodelsFirst1983holland}.
This model generates random undirected graphs with $c$ classes with a label $y_v$ for each node.
A stochastic block matrix $P \in (0, 1)^{c \times c}$ defines the probability $p_{ij}$ that an edge between a pair of nodes with class $i$ and $j$ exists. \\
We use the SBM to generate graphs with ten classes and 100 nodes per class.
For each class, we sample 16-dimensional node features from a Normal distribution centred around different corners of a unit hypercube with standard deviation $\sigma=0.4$.
Edge probabilities $p_{ii}$ (between nodes with the same class) and $p_{ij}$ (between nodes in different classes) are reported in Fig.~\ref{fig:random evaluation}.
The generated graphs used for results in the top left figure ($p_{ii} > p_{ij}$) exhibit strong cluster patterns, while graphs in the bottom right figure ($p_{ii} = p_{ij}$) exhibit no cluster patterns. \\
For graphs with strong cluster patterns, a GCN can achieve high accuracy in a node classification task based on the graph topology alone, while node features play a subordinate role.
On the contrary for graphs with no strong cluster patterns, the features of nodes play a more pronounced role. 
This should highlight the influence of self-loops on the performance of GCNs with different numbers of layers.
Based on our theoretical findings, we thus expect that the performance difference of GCNs with and without self-loops shows a dependency on the number of layers $k$, where we expect models without self-loops to outperform those with self-loops for $k=2$ layers, while we expect the opposite behaviour for $k=1$. 

The GCNs are trained with $80\%$ of the nodes for 70 epochs using an Adam optimizer with learning rate $0.01$.
Fig.~\ref{fig:random evaluation} shows the accuracies of the trained GCNs on the remaining nodes.
We test different numbers of layers $k \in \{1, \dots, 6\}$ for the GCN architecture and average the results over 50 runs.
The code of our experiments is available on GitHub\footnote{\url{https://github.com/M-Lampert/self-loop-paradox}}. \\
The results show that the difference between the accuracies of GCNs with and without self-loops varies with the number of layers $k$.
As expected, with increasing ``noise'' in the topology, i.e. less pronounced cluster patterns, the general performance of a GCN decreases, eventually achieving a performance that is lower than that of an MLP that exclusively uses node features (grey line in Fig.~\ref{fig:random evaluation}).
Moreover, the results suggest that --for graphs with a high level of topological noise-- the performance of a GCN strongly depends on the question of whether the number of layers is odd or even.
In line with our theoretical analysis, our results suggest that the inclusion of self-loops for $k=2$ layers squashes the nodes' own features.
For graphs with the highest level of ``noise'' self-loops actually harm the performance of a two-layered GCN architecture as opposed to a single-layer architecture. \\
In Appendix \ref{app:real-world examples}, we further test our theoretical findings using 23 empirical benchmark graphs from PyTorch Geometric \cite{FastGraphRepresentation2019fey}.
The results in Tab.~\ref{tab:real-world self loop statistics} show that the self-loop paradox indeed holds for 13 of the 23 investigated datasets, which shows the practical relevance of our results.
We also find examples that are not in line with our theoretical prediction, which is likely due to additional correlations (e.g.\ degree-degree correlations) that can invalidate theoretical predictions derived from degree-based ensembles.
We finally investigate the potential impact of the self-loop paradox on the accuracy of a Graph Convolutional Network in a node classification task.
In a first step, we tested the influence of self-loops on accuracy, finding that self-loops are useful for node classification in 15 of the 23 graphs (details in Appendix \ref{app:real-world examples}).
Focusing on those 15 graphs, in Fig.~\ref{fig:real-world examples}, we investigate the performance of GCNs with and without self-loops for different numbers of layers.
The results show that the parity of the number of layers influences the performance of GCNs.
In line with our analytical results, we find that the accuracy increase of a two-layer compared to a one-layer GCN is larger without self-loops than with self-loops for 11 of the 15 graphs.
We suspect that this is due to the self-loop paradox, which states that the inclusion of self-loops limits the backflow of information from a node to itself for two layers.

\section{Conclusion}

Many GNNs add self-loops to the input graph to include feature information about a node itself at each layer.
We prove that this inclusion of self-loops has a counter-intuitive effect, namely that it can decrease the information a node retains about itself in GNNs with two layers.
We analytically prove this finding for random graphs with arbitrary degree sequences.
Experimental validation using a node classification task in synthetically generated graphs shows that our theoretical insights hold for synthetic graphs with ``noisy'' topologies. \\
Our work shows how analytical approaches for the study of statistical ensembles of graphs with given degree sequences --which are frequently used in network science and complex systems theory-- can be applied in the context of GNNs.
In future work, we plan to generalise our analytical proof of the self-loop paradox for arbitrary numbers of layers $k$.
We will further investigate how this effect is influenced by other properties of the graph topology, e.g.\ the average degree, the clustering coefficient, or degree assortativity.
Another promising research direction is to improve our understanding of how the self-loop paradox may influence open challenges in GNNs like over-squashing.

\section*{Acknowledgements}

Moritz Lampert acknowledges funding from the German Federal Ministry of Education and Research, Grant No. 100582863 (TissueNet). Ingo Scholtes acknowledges funding through the Swiss National Science Foundation, Grant No.\ 176938.

\printbibliography

\newpage 

\appendix

\section{Proof of Lemma 2}
\label{sec:lemma2proof}

\begin{proof}
    The expected total number of incoming walks with a length of $2$ is given by
    \begin{align} 
        \mathbb{E}\left(\sum_{u \in V} A^{2}_{uv}\right) = \langle d \rangle \cdot \langle d_N \rangle^{2-1} = \langle d \rangle \cdot \frac{\langle d^2 \rangle}{\langle d \rangle} = \langle d^2 \rangle. 
    \end{align}
    $\tilde{\mathcal{G}}$ only differs from $\mathcal{G}$ in terms of the self-loops which increases both expected degrees by one leading to
    \begin{align}
        \mathbb{E}\left(\sum_{u \in V} \tilde{A}^{2}_{uv}\right) = \left(\langle d \rangle + 1\right) \cdot \left(\langle d_N \rangle + 1\right) = \langle d^2 \rangle + \langle d \rangle + \frac{\langle d^2 \rangle}{\langle d \rangle} + 1.
    \end{align}
    The only closed walks that traverse two edges are those that use the same edge twice. Introducing self-loops only adds one additional option which is traversing the self-loop twice. This means the expected values are given by $\mathbb{E}\left(A^{2}_{vv}\right) = \langle d \rangle$ and $\mathbb{E}\left(\tilde{A}^{2}_{vv}\right) = \langle d \rangle + 1$.
    Putting everything together yields:
    \begin{align}%
        && \frac{\mathbb{E}\left(A^{2}_{vv}\right)}{\mathbb{E}\left(\sum_{u \in V} A^{2}_{uv}\right)} & > \frac{\mathbb{E}\left(\tilde{A}^{2}_{vv}\right)}{\mathbb{E}\left(\sum_{u \in V} \tilde{A}^{2}_{uv}\right)}  \\
        &\Leftrightarrow& \frac{\langle d \rangle}{\langle d^2 \rangle}                                                                                                    & > \frac{\langle d \rangle + 1}{\langle d^2 \rangle + \langle d \rangle + \frac{\langle d^2 \rangle}{\langle d \rangle} + 1} \label{equ:dataset_2nd_layer_estimates} \\
        &\Leftrightarrow& \langle d \rangle^2 + \langle d \rangle                                                                                                          & > 0
    \end{align}%
    This statement is true since $\mathcal{G}$ and $\tilde{\mathcal{G}}$ are connected.
\end{proof}%
The estimates used in the proof and the implications of the lemma are validated empirically in Appendix \ref{app:walk statistics}.

\section{Walk Statistics}\label{app:walk statistics}

Tab.~\ref{tab:self-loop statistics} shows the average proportion of closed walks of length $k$ from a node $v$ to itself out of all walks of length $k$ ending in $v$ for $k \in \{1, 2\}$. 
It also includes the estimates that are used in Equation (\ref{equ:dataset_2nd_layer_estimates}) of the proof.
The statistics show that the proportion of closed walks of length $2$ is smaller in the graph with self-loops than in the graph without self-loops which validates the theoretical findings.
Furthermore, since all of the estimates are in the range of the standard deviation of the averaged proportions, it is reasonable to assume that the estimates are correct.
Note that the estimates for $k=1$ are $0$ without self-loops since there is no other closed walk of length $1$ than a self-loop. With self-loops, the estimate is $\nicefrac{1}{\langle d \rangle + 1}$ since there is one walk from or to each neighbour and the self-loop.
\begin{table}[!ht]%
    \sisetup{round-mode=places, round-precision=3, round-pad = false, scientific-notation=fixed, fixed-exponent=0, mode = text, reset-text-series = false}%
    \centering%
        \begin{tabular}{ll|cccc}%
            \toprule
            \textbf{Dataset} & \textbf{$\circlearrowleft$} & \textbf{$k=1$} & \textbf{$\mathbb{E}_{k=1}$} & \textbf{$k=2$} & \textbf{$\mathbb{E}_{k=2}$} \\
            \midrule
            \multirow[t]{2}{*}{$p_{ii}=\nicefrac{6}{100}$, $p_{ij}=\nicefrac{3}{900}$} & \tikzcmark & \textbf{\num{0.11343789413201177}\textpm \num{0.04374852521544722}} & \textbf{\num{0.10187449078590886}} & \num{0.0955176382372851}\textpm \num{0.010699700021302787} & \num{0.09262450267237515} \\
            & \tikzxmark & \num{0.0}\textpm \num{0.0} & \num{0.0} & \textbf{\num{0.10321466}}\textpm \num{0.01172149} & \textbf{\num{0.10207957}} \\
           \cline{1-6}
           \multirow[t]{2}{*}{$p_{ii}=\nicefrac{6}{100}$, $p_{ij}=\nicefrac{6}{900}$} & \tikzcmark & \textbf{\num{0.08419332348982436}\textpm \num{0.026486313317534156}} & \textbf{\num{0.07774840334445904}} & \num{0.07335720425126081}\textpm \num{0.005401147796338836} & \num{0.07222005956635401} \\
            & \tikzxmark & \num{0.0}\textpm \num{0.0} & \num{0.0} & \textbf{\num{0.078173354}}\textpm \num{0.0062217484} & \textbf{\num{0.077841796}} \\
           \cline{1-6}
           \multirow[t]{2}{*}{$p_{ii}=\nicefrac{3}{100}$, $p_{ij}=\nicefrac{6}{900}$} & \tikzcmark & \textbf{\num{0.11582836097081453}\textpm \num{0.04478672022670924}} & \textbf{\num{0.10322048186935921}} & \num{0.09605386066618261}\textpm \num{0.010502718346931868} & \num{0.09289395453229249} \\
            & \tikzxmark & \num{0.0}\textpm \num{0.0} & \num{0.0} & \textbf{\num{0.10366025}}\textpm \num{0.012137852} & \textbf{\num{0.10240694}} \\
           \cline{1-6}
           \multirow[t]{2}{*}{$p_{ii}=\nicefrac{1}{100}$, $p_{ij}=\nicefrac{9}{900}$} & \tikzcmark & \textbf{\num{0.09800245758283682}\textpm \num{0.031533340705481964}} & \textbf{\num{0.08983111774536724}} & \num{0.08433406379427907}\textpm \num{0.007093750851698074} & \num{0.08261578854855903} \\
            & \tikzxmark & \num{0.0}\textpm \num{0.0} & \num{0.0} & \textbf{\num{0.090621635}}\textpm \num{0.008397878} & \textbf{\num{0.090055816}} \\
            \bottomrule
        \end{tabular}%
    \caption{The number of closed walks with length $k$ for each node $v$ divided by the total number of walks ending in $v$ and the corresponding estimates $\mathbb{E}_{k=i}$ from Equation (\ref{equ:dataset_2nd_layer_estimates}) of the proof for $k\in\{1,2\}$. The proportions are averaged over all nodes and the standard deviation is reported. The larger value is printed in bold for each graph and walk length $k$ and their corresponding pair of proportions with and without self-loops. Additionally, the standard deviation is printed in bold if the ranges of both standard deviations do not overlap.}\label{tab:self-loop statistics}%
\end{table}%

The statistics in Tab.~\ref{tab:more self-loop statistics} report the average walk proportions for larger $k$ and suggest that similar behaviour can be observed for all even $k$.
\begin{table}[!ht]%
    \sisetup{round-mode=places, round-precision=3, round-pad = false, scientific-notation=fixed, fixed-exponent=0, mode = text, reset-text-series = false}%
    \centering%
    \begin{tabular}{ll|cccc}%
        \toprule
        \textbf{Dataset} & \textbf{$\circlearrowleft$} & \textbf{$k=3$} & \textbf{$k=4$} & \textbf{$k=5$} & \textbf{$k=6$} \\
        \midrule
        \multirow[t]{2}{*}{$p_{ii}=\nicefrac{6}{100}$, $p_{ij}=\nicefrac{3}{900}$} & \tikzcmark & \textbf{\num{0.02660338419075669}\textpm \num{0.003550903051330844}} & \num{0.018959660493627566}\textpm \num{0.0026638962288292756} & \textbf{\num{0.008475457974386473}\textpm \num{0.0016680484271446015}} & \num{0.005931625931971191}\textpm \num{0.0014702456086939366} \\
        & \tikzxmark & \num{0.0022692676}\textpm \num{0.0023886662} & \textbf{\num{0.020353615}}\textpm \num{0.0034239846} & \num{0.0020546913}\textpm \num{0.0012198763} & \textbf{\num{0.005976999}}\textpm \num{0.0015999848} \\
        \cline{1-6}
        \multirow[t]{2}{*}{$p_{ii}=\nicefrac{6}{100}$, $p_{ij}=\nicefrac{6}{900}$} & \tikzcmark & \textbf{\num{0.016272000621432647}\textpm \num{0.0014893189723001253}} & \num{0.011521913974739206}\textpm \num{0.001466775170838532} & \textbf{\num{0.00445478580877681}\textpm \num{0.0008168794597159306}} & \num{0.0031159655550918658}\textpm \num{0.0007160263346802578} \\
        & \tikzxmark & \num{0.0014473675}\textpm \num{0.0013037672} & \textbf{\num{0.012192862}}\textpm \num{0.0018050302} & \num{0.0012805075}\textpm \num{0.00056902354} & \textbf{\num{0.0031241702}}\textpm \num{0.00075817463} \\
        \cline{1-6}
        \multirow[t]{2}{*}{$p_{ii}=\nicefrac{3}{100}$, $p_{ij}=\nicefrac{6}{900}$} & \tikzcmark & \textbf{\num{0.02569805025383052}\textpm \num{0.0032012589738738763}} & \num{0.018027546544119928}\textpm \num{0.002377286659590112} & \textbf{\num{0.007464153316087125}\textpm \num{0.0012527208619002332}} & \num{0.005078400959528008}\textpm \num{0.001156148762275348} \\
        & \tikzxmark & \num{0.0008239672}\textpm \num{0.0013737144} & \textbf{\num{0.01952516}}\textpm \num{0.00322158} & \num{0.00097001524}\textpm \num{0.0006396858} & \textbf{\num{0.005268694}}\textpm \num{0.0013791666} \\
        \cline{1-6}
        \multirow[t]{2}{*}{$p_{ii}=\nicefrac{1}{100}$, $p_{ij}=\nicefrac{9}{900}$} & \tikzcmark & \textbf{\num{0.020241689277317294}\textpm \num{0.001998253799888017}} & \num{0.014346792411064064}\textpm \num{0.0017691784430998963} & \textbf{\num{0.005571283053296202}\textpm \num{0.0008980889078624852}} & \num{0.003841105397054509}\textpm \num{0.0008391491777436528} \\
        & \tikzxmark & \num{0.0006421652}\textpm \num{0.0010376003} & \textbf{\num{0.015488376}}\textpm \num{0.0023351228} & \num{0.0008891595}\textpm \num{0.0004677999} & \textbf{\num{0.0039831055}}\textpm \num{0.00097068737} \\
    \bottomrule
    \end{tabular}%
    \caption{The average proportions as in Tab.~\ref{tab:self-loop statistics} for larger $k$.}\label{tab:more self-loop statistics}%
\end{table}%

\section{Real-World Examples}\label{app:real-world examples}

To investigate this effect on real-world data, we use 23 datasets from PyTorch Geometric \cite{FastGraphRepresentation2019fey}. 
Since we have only considered undirected graphs in our theoretical analysis, we drop the direction of the edges in the datasets that are directed.
To avoid mathematical errors, we remove isolated nodes from all datasets.
We further disregard existing self-loops and either remove them or add the missing ones. 

Tab.~\ref{tab:real-world self loop statistics} shows the statistics discussed in Appendix \ref{app:walk statistics} for the real-world examples.
We can see that the degrees vary far more in the real-world examples than in the synthetic graphs.
Consequently, the standard deviations are larger and the estimates are less accurate.
The expected pattern from our theoretical findings that the proportions are larger for walks of length 2 without self-loops cannot be observed in all of the datasets.
Although the estimates falsely predict the pattern in all datasets, they still give a good indication of when the pattern exists because the difference is very small ($<0.001$) in most datasets where the pattern does not exist.
The predicted pattern also continues for larger walk lengths for some of the datasets which further supports our hypothesis from Appendix \ref{app:walk statistics} that this alternating pattern continues for larger walk lengths.

\begin{table}[!hp]%
    \sisetup{round-mode=places, round-precision=3, round-pad = false, scientific-notation=fixed, fixed-exponent=0, mode = text, reset-text-series = false}%
    \centering
    \begin{footnotesize}
        \begin{tabular}{llcccccc}
            \toprule
            \textbf{Dataset} & \textbf{$\circlearrowleft$} & \textbf{$k=1$} & \textbf{$\mathbb{E}_{k=1}$} & \textbf{$k=2$} & \textbf{$\mathbb{E}_{k=2}$} & \textbf{$k=3$} & \textbf{$k=4$} \\
            \midrule
            \textcolor{custom2}{Planetoid:} & \tikzcmark & \textbf{\num{0.2753171986954344}\textpm \num{0.12774072320635851}} & \textbf{\num{0.20416163600959422}} & \num{0.17374195727135344}\textpm \num{0.11240366035757633} & \num{0.08396837164179215} & \textbf{\num{0.09225183174779077}}\textpm \num{0.1062133935119413} & \num{0.0689790135700717}\textpm \num{0.10464343981486578} \\
            \textcolor{custom2}{Cora} & \tikzxmark & \num{0.0}\textpm \num{0.0} & \num{0.0} & \textbf{\num{0.21260795}}\textpm \num{0.20733474} & \textbf{\num{0.091665365}} & \num{0.016834421}\textpm \num{0.031987876} & \textbf{\num{0.090490595}}\textpm \num{0.20378947} \\
            \cline{1-8}
            \textcolor{custom2}{Planetoid:} & \tikzcmark & \textbf{\num{0.3500960933022648}\textpm \num{0.14010157630742734}} & \textbf{\num{0.2647985224379017}} & \num{0.27402875867598203}\textpm \num{0.13570477884207083} & \num{0.12636722031701778} & \textbf{\num{0.1995069101375161}}\textpm \num{0.1573648608020989} & \num{0.17616824342669976}\textpm \num{0.16411067660410725} \\
            \textcolor{custom2}{CiteSeer} & \tikzxmark & \num{0.0}\textpm \num{0.0} & \num{0.0} & \textbf{\num{0.3988276}}\textpm \num{0.31116065} & \textbf{\num{0.14464569}} & \num{0.01707225}\textpm \num{0.039937202} & \textbf{\num{0.28699535}}\textpm \num{0.35043353} \\
            \cline{1-8}
            Planetoid: & \tikzcmark & \textbf{\num{0.33947667546004556}\textpm \num{0.1687293370083133}} & \textbf{\num{0.18194988420514444}} & \textbf{\num{0.10440860540592067}}\textpm \num{0.06799267994146202} & \num{0.056249447832670364} & \textbf{\num{0.03257595628999932}}\textpm \num{0.03748430825288423} & \textbf{\num{0.017871887051941957}}\textpm \num{0.024460339874375406} \\
            PubMed & \tikzxmark & \num{0.0}\textpm \num{0.0} & \num{0.0} & \num{0.091916986}\textpm \num{0.09430054} & \textbf{\num{0.05960203}} & \num{0.0016883755}\textpm \num{0.0058518667} & \num{0.017526256}\textpm \num{0.03066294} \\
            \cline{1-8}
            \textcolor{custom2}{CitationFull:} & \tikzcmark & \textbf{\num{0.27184352573053755}\textpm \num{0.16523902873115354}} & \textbf{\num{0.14350748898093596}} & \num{0.151854903788674}\textpm \num{0.12828143723625166} & \num{0.046267213502470865} & \textbf{\num{0.07621134537515015}}\textpm \num{0.12017481905731667} & \num{0.057391816916176494}\textpm \num{0.11733788167410086} \\
            \textcolor{custom2}{DBLP} & \tikzxmark & \num{0.0}\textpm \num{0.0} & \num{0.0} & \textbf{\num{0.18573442}}\textpm \num{0.23463286} & \textbf{\num{0.048511717}} & \num{0.004685337}\textpm \num{0.012634584} & \textbf{\num{0.08913916}}\textpm \num{0.2331492} \\
            \cline{1-8}
            \textcolor{custom2}{Amazon:} & \tikzcmark & \textbf{\num{0.0788756246598174}}\textpm \num{0.09644657409924859} & \textbf{\num{0.026665055435899972}} & \num{0.02235594292286273}\textpm \num{0.0449194046445746} & \num{0.00571448160627339} & \textbf{\num{0.0055792725663614436}}\textpm \num{0.03338061592652591} & \num{0.003590332367882722}\textpm \num{0.031680452865369656} \\
            \textcolor{custom2}{Computers} & \tikzxmark & \num{0.0}\textpm \num{0.0} & \num{0.0} & \textbf{\num{0.023424393}}\textpm \num{0.06764839} & \textbf{\num{0.005747325}} & \num{0.0023119026}\textpm \num{0.012156788} & \textbf{\num{0.005118413}}\textpm \num{0.057773616} \\
            \cline{1-8}
            \textcolor{custom2}{Amazon:} & \tikzcmark & \textbf{\num{0.08064094300922788}}\textpm \num{0.09587741911789843} & \textbf{\num{0.03066785566065355}} & \num{0.026096422129453397}\textpm \num{0.04479338671110393} & \num{0.009622897560427756} & \textbf{\num{0.007877852626339927}}\textpm \num{0.035428929847985756} & \num{0.005579726484124402}\textpm \num{0.034426237878067374} \\
            \textcolor{custom2}{Photo} & \tikzxmark & \num{0.0}\textpm \num{0.0} & \num{0.0} & \textbf{\num{0.027591048}}\textpm \num{0.072085425} & \textbf{\num{0.009716397}} & \num{0.004052574}\textpm \num{0.012209515} & \textbf{\num{0.0074484884}}\textpm \num{0.06463354} \\
            \cline{1-8}
            \textcolor{custom2}{Coauthor:} & \tikzcmark & \textbf{\num{0.16983277704112135}\textpm \num{0.11850693871667325}} & \textbf{\num{0.10066384361130924}} & \num{0.09023774890279579}\textpm \num{0.0636753946683909} & \num{0.05203738927240903} & \textbf{\num{0.02979417715637343}}\textpm \num{0.036273544115593866} & \textbf{\num{0.014179928826730578}}\textpm \num{0.02239186248450821} \\
            \textcolor{custom2}{CS} & \tikzxmark & \num{0.0}\textpm \num{0.0} & \num{0.0} & \textbf{\num{0.09218002}}\textpm \num{0.07681128} & \textbf{\num{0.054893926}} & \num{0.010777668}\textpm \num{0.015071858} & \num{0.011500124}\textpm \num{0.02056862} \\
            \cline{1-8}
            \textcolor{custom2}{Wikipedia:} & \tikzcmark & \textbf{\num{0.11458248381172421}}\textpm \num{0.11820033344900575} & \textbf{\num{0.03502053153205808}} & \num{0.028410243790987885}\textpm \num{0.04880239043570704} & \num{0.009380244984107389} & \textbf{\num{0.009624134924088834}}\textpm \num{0.023728623088808624} & \textbf{\num{0.006285199527719737}}\textpm \num{0.013120278037002444} \\
            \textcolor{custom2}{Chameleon} & \tikzxmark & \num{0.0}\textpm \num{0.0} & \num{0.0} & \textbf{\num{0.028610826}}\textpm \num{0.05530472} & \textbf{\num{0.009469067}} & \num{0.005133608}\textpm \num{0.011901251} & \num{0.00582118}\textpm \num{0.011708718} \\
            \cline{1-8}
            Wikipedia: & \tikzcmark & \textbf{\num{0.09086328100425137}}\textpm \num{0.10702455420286726} & \textbf{\num{0.012940804546974098}} & \textbf{\num{0.01951224452912546}}\textpm \num{0.036020446700296964} & \num{0.0023867849827599406} & \textbf{\num{0.003362435781398157}}\textpm \num{0.010472713155583872} & \textbf{\num{0.0011859082395043272}}\textpm \num{0.003914279371318472} \\
            Squirrel & \tikzxmark & \num{0.0}\textpm \num{0.0} & \num{0.0} & \num{0.018372118}\textpm \num{0.035970498} & \textbf{\num{0.0023924955}} & \num{0.0012673845}\textpm \num{0.0035898502} & \num{0.0010557345}\textpm \num{0.0036361383} \\
            \cline{1-8}
            \multirow[t]{2}{*}{Actor} & \tikzcmark & \textbf{\num{0.24773061137507307}\textpm \num{0.15880094679095672}} & \textbf{\num{0.1247578715648637}} & \textbf{\num{0.08084478235729907}}\textpm \num{0.08051209970640501} & \num{0.018854846147781514} & \textbf{\num{0.01524848158220161}}\textpm \num{0.032462243812594986} & \textbf{\num{0.00455549036466544}}\textpm \num{0.012755459513039363} \\
                & \tikzxmark & \num{0.0}\textpm \num{0.0} & \num{0.0} & \num{0.07652268}\textpm \num{0.08982736} & \textbf{\num{0.019217184}} & \num{0.0010713629}\textpm \num{0.0054779933} & \num{0.0034055777}\textpm \num{0.010085894} \\
            \cline{1-8}
            Airports: & \tikzcmark & \textbf{\num{0.1959901237274291}\textpm \num{0.1727043662551911}} & \textbf{\num{0.04191911927377761}} & \textbf{\num{0.030099259242845386}}\textpm \num{0.05039075643783647} & \num{0.010477485502397497} & \textbf{\num{0.0066706280327052}}\textpm \num{0.031320317715542945} & \num{0.003948154109802572}\textpm \num{0.029320086506934576} \\
            USA & \tikzxmark & \num{0.0}\textpm \num{0.0} & \num{0.0} & \num{0.027469004}\textpm \num{0.07061928} & \textbf{\num{0.010588425}} & \num{0.0025302158}\textpm \num{0.0056702974} & \textbf{\num{0.005394706}}\textpm \num{0.05793205} \\
            \cline{1-8}
            Airports: & \tikzcmark & \textbf{\num{0.1127968389157385}}\textpm \num{0.12584871081947843} & \textbf{\num{0.03221639095408417}} & \textbf{\num{0.019573940995356906}}\textpm \num{0.0232626164923049} & \num{0.014079146039569697} & \textbf{\num{0.003585590264416338}}\textpm \num{0.0029240770163303244} & \textbf{\num{0.002877351629693407}}\textpm \num{0.0025988687535484292} \\
            Europe & \tikzxmark & \num{0.0}\textpm \num{0.0} & \num{0.0} & \num{0.01730689}\textpm \num{0.01938114} & \textbf{\num{0.014280199}} & \num{0.0027309733}\textpm \num{0.0024715774} & \num{0.00285833}\textpm \num{0.002624297} \\
            \cline{1-8}
            Airports: & \tikzcmark & \textbf{\num{0.1596005120638816}\textpm \num{0.1510973747088236}} & \textbf{\num{0.06130089008170287}} & \textbf{\num{0.032607481488784}}\textpm \num{0.0121851130077224} & \num{0.02902954888151271} & \textbf{\num{0.010358520678639902}}\textpm \num{0.00664895507470886} & \textbf{\num{0.008550160860850503}}\textpm \num{0.007132599827304364} \\
            Brazil & \tikzxmark & \num{0.0}\textpm \num{0.0} & \num{0.0} & \num{0.028981725}\textpm \num{0.012137854} & \textbf{\num{0.029897459}} & \num{0.008120432}\textpm \num{0.007033595} & \num{0.0084477905}\textpm \num{0.007219271} \\
            \cline{1-8}
            Twitch: & \tikzcmark & \textbf{\num{0.10749043048967134}}\textpm \num{0.11997786167700288} & \textbf{\num{0.030078473165004768}} & \textbf{\num{0.011045386733080283}}\textpm \num{0.02735128621975906} & \num{0.0039026441744488185} & \textbf{\num{0.0004648020624712894}}\textpm \num{0.002692635586394934} & \textbf{\num{0.0001835239020902839}}\textpm \num{0.00023319108695959049} \\
            DE & \tikzxmark & \num{0.0}\textpm \num{0.0} & \num{0.0} & \num{0.009455832}\textpm \num{0.024991876} & \textbf{\num{0.0039179344}} & \num{0.00016511849}\textpm \num{0.00020588843} & \num{0.00017953952}\textpm \num{0.00022455867} \\
            \cline{1-8}
            Twitch: & \tikzcmark & \textbf{\num{0.2171501218959997}\textpm \num{0.15558566433970294}} & \textbf{\num{0.09162445417060276}} & \textbf{\num{0.05104298075469857}}\textpm \num{0.06845865783786001} & \num{0.01650860062476595} & \textbf{\num{0.006728277680000525}}\textpm \num{0.020952626554511633} & \textbf{\num{0.0017404075563294446}}\textpm \num{0.0064757725686546055} \\
            EN & \tikzxmark & \num{0.0}\textpm \num{0.0} & \num{0.0} & \num{0.046679903}\textpm \num{0.07300203} & \textbf{\num{0.01678571}} & \num{0.0008070135}\textpm \num{0.0028376034} & \num{0.0013402}\textpm \num{0.0042007165} \\
            \cline{1-8}
            Twitch: & \tikzcmark & \textbf{\num{0.1627429067915542}\textpm \num{0.14407387419371886}} & \textbf{\num{0.05551124506959817}} & \textbf{\num{0.025490183595571942}}\textpm \num{0.046102539435202365} & \num{0.008325968347750262} & \textbf{\num{0.0019564718809174557}}\textpm \num{0.008755151700990128} & \textbf{\num{0.0006426729755525367}}\textpm \num{0.0034665004525513497} \\
            RU & \tikzxmark & \num{0.0}\textpm \num{0.0} & \num{0.0} & \num{0.022015173}\textpm \num{0.04395512} & \textbf{\num{0.008395872}} & \num{0.0003856454}\textpm \num{0.0007395304} & \num{0.00058043335}\textpm \num{0.00230696} \\
            \cline{1-8}
            \textcolor{custom2}{WebKB:} & \tikzcmark & \textbf{\num{0.35027673090658423}\textpm \num{0.1405041156461585}} & \textbf{\num{0.24830394177913634}} & \num{0.1322082128073393}\textpm \num{0.11874229760223443} & \num{0.04927071911534715} & \textbf{\num{0.04554635676283532}}\textpm \num{0.05263955464856867} & \textbf{\num{0.024841605966097073}}\textpm \num{0.024467273547213727} \\
            \textcolor{custom2}{Cornell} & \tikzxmark & \num{0.0}\textpm \num{0.0} & \num{0.0} & \textbf{\num{0.14292945}}\textpm \num{0.15008567} & \textbf{\num{0.05182413}} & \num{0.006088204}\textpm \num{0.011886508} & \num{0.023082197}\textpm \num{0.030792281} \\
            \cline{1-8}
            \textcolor{custom2}{WebKB:} & \tikzcmark & \textbf{\num{0.3527322404371585}\textpm \num{0.13264052422243996}} & \textbf{\num{0.2469635663287876}} & \num{0.1179449233747085}\textpm \num{0.11375329122248447} & \num{0.04162936185952694} & \textbf{\num{0.03547148078469452}}\textpm \num{0.04233541382800199} & \textbf{\num{0.01822495571997832}}\textpm \num{0.018385903516872373} \\
            \textcolor{custom2}{Texas} & \tikzxmark & \num{0.0}\textpm \num{0.0} & \num{0.0} & \textbf{\num{0.121115334}}\textpm \num{0.14219332} & \textbf{\num{0.043437645}} & \num{0.0032399425}\textpm \num{0.0059923055} & \num{0.0163292}\textpm \num{0.021792298} \\
            \cline{1-8}
            \textcolor{custom2}{WebKB:} & \tikzcmark & \textbf{\num{0.3103389998237438}\textpm \num{0.13328838452187236}} & \textbf{\num{0.21807124298514255}} & \num{0.11865200107883739}\textpm \num{0.10647525365444252} & \num{0.04522158679594363} & \textbf{\num{0.03555723456896294}}\textpm \num{0.04434079023021657} & \textbf{\num{0.0181733135913184}}\textpm \num{0.01860935997090211} \\
            \textcolor{custom2}{Wisconsin} & \tikzxmark & \num{0.0}\textpm \num{0.0} & \num{0.0} & \textbf{\num{0.12335342}}\textpm \num{0.12938763} & \textbf{\num{0.047363438}} & \num{0.0041821375}\textpm \num{0.008422742} & \num{0.016383175}\textpm \num{0.019072773} \\
            \cline{1-8}
            LastFM- & \tikzcmark & \textbf{\num{0.25062682775442036}\textpm \num{0.16323281594802463}} & \textbf{\num{0.12056423883487621}} & \textbf{\num{0.09755182817298726}}\textpm \num{0.08574471031293157} & \num{0.03784715098093033} & \textbf{\num{0.025993335133682505}}\textpm \num{0.040866427417655705} & \textbf{\num{0.010763609765008484}}\textpm \num{0.021633415985961774} \\
            Asia & \tikzxmark & \num{0.0}\textpm \num{0.0} & \num{0.0} & \num{0.09383895}\textpm \num{0.100823306} & \textbf{\num{0.039335903}} & \num{0.0043833344}\textpm \num{0.008357784} & \num{0.008732138}\textpm \num{0.021267174} \\
            \cline{1-8}
            \textcolor{custom2}{Amazon-} & \tikzcmark & \textbf{\num{0.13717246494243585}\textpm \num{0.038715520579149384}} & \textbf{\num{0.11630071198937941}} & \num{0.0935051500191588}\textpm \num{0.03470600531090694} & \num{0.07523946358934275} & \textbf{\num{0.056583720462568314}}\textpm \num{0.03354131307760757} & \textbf{\num{0.04995491661717454}}\textpm \num{0.0317915308355309} \\
            \textcolor{custom2}{Ratings} & \tikzxmark & \num{0.0}\textpm \num{0.0} & \num{0.0} & \textbf{\num{0.101498984}}\textpm \num{0.04404837} & \textbf{\num{0.081361026}} & \num{0.041748278}\textpm \num{0.031576995} & \num{0.049615096}\textpm \num{0.03161969} \\
            \cline{1-8}
            \textcolor{custom2}{Roman-} & \tikzcmark & \textbf{\num{0.2708345606677616}\textpm \num{0.058951858373089076}} & \textbf{\num{0.25602151743344154}} & \num{0.24624684796798452}\textpm \num{0.029787532993463817} & \num{0.2343058285885061} & \textbf{\num{0.1735682411644773}\textpm \num{0.030147319133717945}} & \num{0.1492730277368328}\textpm \num{0.029802465243873325} \\
            \textcolor{custom2}{Empire} & \tikzxmark & \num{0.0}\textpm \num{0.0} & \num{0.0} & \textbf{\num{0.31554937}}\textpm \num{0.05918201} & \textbf{\num{0.30600446}} & \num{0.05881168}\textpm \num{0.043280795} & \textbf{\num{0.17157099}}\textpm \num{0.04335424} \\
            \cline{1-8}
            \textcolor{custom2}{Mine-} & \tikzcmark & \textbf{\num{0.11334444444444448}\textpm \num{0.011122766109467821}} & \textbf{\num{0.11260753790805393}} & \num{0.11256025000000001}\textpm \num{0.005175395698103137} & \num{0.11204955217840339} & \textbf{\num{0.06843759543381403}\textpm \num{0.003931543359049382}} & \textbf{\num{0.0560096748485016}}\textpm \num{0.0031095965109960953} \\
            \textcolor{custom2}{sweeper} & \tikzxmark & \num{0.0}\textpm \num{0.0} & \num{0.0} & \textbf{\num{0.1267052}\textpm \num{0.0059290915}} & \textbf{\num{0.12618896}} & \num{0.047669828}\textpm \num{0.002691617} & \num{0.053629268}\textpm \num{0.002950947} \\
            \bottomrule
        \end{tabular}
    \end{footnotesize}
    \caption{The average proportions and statistics as in Tab.~\ref{tab:self-loop statistics} and Tab.~\ref{tab:more self-loop statistics} for real-world datasets. The names of the datasets that are in line with our theoretical results are marked in orange. As in Tab.~\ref{tab:self-loop statistics}, the larger number of each value pair is printed in bold. Note that some values are rounded to the same values. In this case, the value that was larger before rounding is printed in bold.}\label{tab:real-world self loop statistics}%
\end{table}%

The aforementioned datasets are further used to evaluate the performance of the GCN model on node classification tasks.
The training parameters are set to similar values as above. 
We train the GCNs for 70 epochs on 80\% of the nodes.
An Adam optimizer is used with learning rate $0.01$ and we include a dropout layer with a dropout rate of $0.2$ after each GCN layer.

\begin{figure}[!hp]%
    \centering%
    \includegraphics[width=\textwidth]{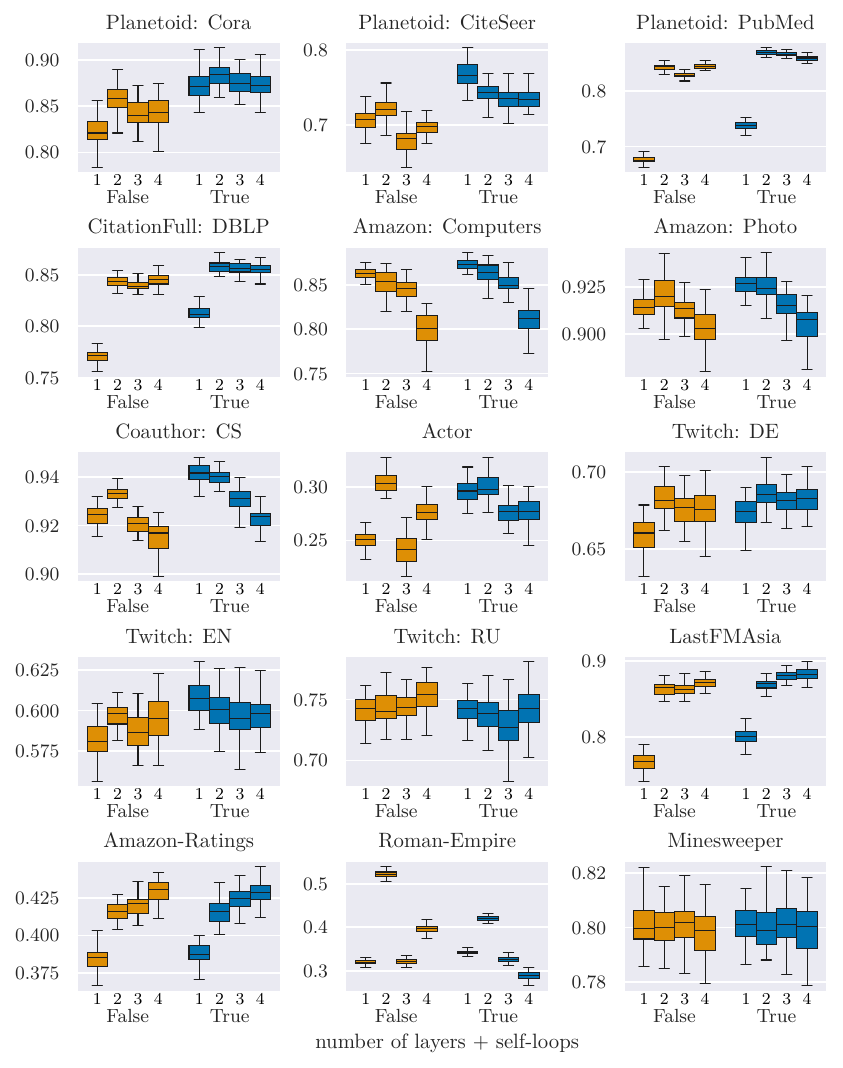}%
    \caption{Test set accuracies as in Fig.~\ref{fig:random evaluation} for real-world examples. Outliers outside of the whiskers are omitted for readability.}\label{fig:real-world examples}%
\end{figure}

Our theoretical results make statements about the amount of information a node retains about itself for MPNN architectures with two layers.
Thus, it is reasonable to assume that we can only expect results that are in line with our theoretical findings if the node's own feature is useful for the node classification task.
We measure the usefulness of a node's own feature on the performance of single-layered GCNs.
The node's own feature does not influence the predictions of single-layered GCNs without self-loops and thus if that GCN performs better than the one with self-loops, the node's own feature can be assumed to be not useful or even harmful to the task.
We can observe those results for 8 of the investigated datasets and report only the remaining 15 datasets in Fig.~\ref{fig:real-world examples}.

The results show that the parity of the number of message passing layers strongly influences the performance of GCNs.
In line with our analytical results, we further find that for 11 out of 15 data sets the accuracy increase of a two-layer compared to a one-layer GCN is larger for an input graph without self-loops than a graph with self-loops.
We suspect that this is due to the self-loop paradox, which states that the inclusion of self-loops actually limits the backflow of information from a node to itself for two message passing layers.

\pagebreak

\section{The Self-Loop Paradox for Other GNNs}\label{app:gnn types}

So far, we only considered the Graph Convolutional Network (GCN) \cite{SemiSupervisedClassificationGraph2017kipf} although there exists a large variety of other GNN architectures. 
Lemma \ref{lem:influence of input feature} holds for all Message Passing GNNs (MPNNs) --sometimes also called Weisfeiler-Leman (WL) test based GNNs \cite{RedundancyFreeMessagePassing2022chen}. 
By definition of \citet{NeuralMessagePassing2017gilmer}, this includes all GNNs that can be expressed with a message function 
\begin{align}
    \mathbf{m}_v^{(k+1)} = \sum_{u \in N(v)} M^{(k)}\left(\mathbf{h}_v^{(k)}, \mathbf{h}_u^{(k)}, \mathbf{e}_{vu}\right)
\end{align}
and an update function
\begin{align}
    \mathbf{h}_v^{(k+1)} = U^{(k)}\left(\mathbf{h}_v^{(k)}, \mathbf{m}_v^{(k+1)}\right).
\end{align}
This includes GNNs like the GCN \cite{SemiSupervisedClassificationGraph2017kipf}, GraphSAGE \cite{InductiveRepresentationLearning2017hamilton} or the Graph Isomorphism Network (GIN) \cite{HowPowerfulAre2019xu} but does not include models like Personalized Propagation of Neural Predictions (PPNP) \cite{PredictThenPropagate2019klicpera}.

In contrast to the original formulation of the GCN, this general framework does not rely on self-loops for all nodes since it includes the node's own feature in the update step. 
Our theoretical findings, on the other hand, are based on the assumption that the node's own feature is included via self-loops.
This means our findings do not apply to all MPNNs but only to those that can be reformulated using a modified message step that includes self-loops 
\begin{align}
    \mathbf{m}_v^{(k+1)} = \sum_{u \in N(v) \cup \{v\}} M^{(k)}\left(\mathbf{h}_v^{(k)}, \mathbf{h}_u^{(k)}, \mathbf{e}_{vu}\right)
\end{align}
and an update step that excludes the node's own feature
\begin{align}
    \mathbf{h}_v^{(k+1)} = U^{(k)}\left(\mathbf{m}_v^{(k+1)}\right).
\end{align}

This includes the GCN as defined in Equation (\ref{equ:gcn layer}) but also the Graph Attention Network (GAT) by \citet{GraphAttentionNetworks2018velickovic}:
\begin{align}
    \mathbf{m}_v^{(k+1)} &= \sum_{u \in N(v) \cup \{v\}} a_{vu}^{(k)} \mathbf{h}^{(k)}_u W^{(k)} \\
    \mathbf{h}_v^{(k+1)} &= \sigma \left(\mathbf{m}_v^{(k+1)} + \mathbf{b}^{(k)}\right)
\end{align}
Technically, this is only true if one disregards that the Softmax function considers the whole neighbourhood $N(v)$ to calculate the attention weights $a_{vu}$ which would require a message function $M^{(k)}\left(\mathbf{h}_v^{(k)}, \mathbf{h}_u^{(k)}, \mathbf{e}_{vu}, \{h_w^{(k)}: w\in N(v)\}\right)$.
Other models like GraphSAGE and GIN do not fall into this category since they both either concatenate or add the node's feature $\mathbf{h}_v^{(k)}$ to the aggregated message $\mathbf{m}_v^{(k+1)}$ in the update function.

\end{document}